\documentclass[conference]{IEEEtran}
\IEEEoverridecommandlockouts

\usepackage{cite}
\usepackage{amsmath,amssymb,amsfonts}
\usepackage{algorithmic}
\usepackage{graphicx}
\usepackage{textcomp}
\usepackage{xcolor}
\usepackage{tcolorbox}
\usepackage{multicol}
\usepackage{listings}
\usepackage{xcolor}  
\usepackage{algorithm}
\usepackage{algorithmic}
\usepackage{pgfplots}
\usepackage{url} 
\usepackage{multirow}
\pgfplotsset{compat=1.17}
\usepackage{xcolor}

\usepackage{listings}
\usepackage{tcolorbox}
\tcbuselibrary{listings, breakable, skins}
\definecolor{vividorange}{RGB}{255,100,20}

\usepackage{amssymb}
\usepackage{pifont}
\usepackage{booktabs}

\newcommand{\cmark}{\ding{51}} 
\newcommand{\xmark}{\ding{55}} 

\tcbset{
  mylisting/.style={
    enhanced,
    breakable,
    colback=gray!10,       
    colframe=gray!50,      
    listing only,
    listing options={
      basicstyle=\ttfamily\small,
      breaklines=true,
      numbers=none,
      xleftmargin=2mm,
      xrightmargin=2mm
    },
    boxrule=0.5pt,
    arc=2pt,
    outer arc=2pt,
    top=1mm,
    bottom=1mm
  }
}

\def\BibTeX{{\rm B\kern-.05em{\sc i\kern-.025em b}\kern-.08em
    T\kern-.1667em\lower.7ex\hbox{E}\kern-.125emX}}
\begin{document}

\title{Incorporating Contextual Paralinguistic Understanding in Large Speech-Language Models\\
\thanks{
$^{*}$These authors contributed equally to this work.}
\thanks{
This research/project is supported by the National Research Foundation, Singapore, under its National Large Language Models Funding Initiative. Any opinions, findings, conclusions or recommendations expressed in this material are those of the author(s) and do not reflect the views of the National Research Foundation, Singapore. }
}

\author{
\IEEEauthorblockN{Qiongqiong Wang$^{*}$, Hardik B. Sailor$^{*}$, Jeremy H. M. Wong, Tianchi Liu, Shuo Sun,  \\Wenyu Zhang, Muhammad Huzaifah, Nancy Chen, Ai Ti Aw }
\IEEEauthorblockA{\textit{Institute for Infocomm Research (I$^2$R)},
\textit{Agency for Science, Technology and Research (A$\star$STAR)}\\
Singapore \\
\{wang\underline{\enskip}qiongqiong, sailor\underline{\enskip}hardik\underline{\enskip}bhupendra\}@i2r.a-star.edu.sg}
}

\maketitle

\begin{abstract}
Current large speech language models (Speech-LLMs) often exhibit limitations in empathetic reasoning, primarily due to the absence of training datasets that integrate both contextual content and paralinguistic cues. In this work, we propose two approaches to incorporate contextual paralinguistic information into model training: (1) an explicit method that provides paralinguistic metadata (e.g., emotion annotations) directly to the LLM, and (2) an implicit method that automatically generates novel training question–answer (QA) pairs using both categorical and dimensional emotion annotations alongside speech transcriptions. Our implicit method boosts performance (LLM-judged) by 38.41\% on a human-annotated QA benchmark, reaching 46.02\% when combined with the explicit approach, showing effectiveness in contextual paralinguistic understanding. We also validate the LLM judge by demonstrating its correlation with classification metrics, providing support for its reliability.


\end{abstract}
\begin{IEEEkeywords}
Speech-LLM, multi-modal, contextual-paralinguistic, emotion, data generation.
\end{IEEEkeywords}

\section{Introduction}

In recent years, large language models (LLMs) have shown remarkable capabilities across a wide range of natural language processing tasks. Building on this success, large speech language models (Speech-LLMs), which extend LLMs with speech inputs, have emerged as a promising direction to enable spoken dialog systems, voice-based assistants, and human-computer interaction~\cite{qwen,qwen2,gpt4,salmonn}. While these models excel at content-related tasks like speech recognition, these models often exhibit limitations in tasks requiring empathetic reasoning or emotional understanding. 

Past efforts to improve paralinguistic understanding for LLM can be grouped into: (1) fine-tuning on labeled emotional data~\cite{Lin24,  kimparalinguistics, kang2024frozen, wang24blsp}, (2) knowledge distillation from paralinguistic teachers~\cite{desta, desta2, wang24blsp}, and (3) translating emotional signals into language prompts~\cite{wu-etal-2025-beyond,CLAP4Emo,xu2024secap,wu2024empower}.
Another line of work focuses on building datasets with text instructions for multimodal Speech-LLMs~\cite{cheng2025voxdialogue,pandey2025sift,gong2023joint}. For instance, \cite{pandey2025sift} generated a large-scale dataset using comprehensive metadata and divers instructions. However, the question-answer (QA) pairs primarily target acoustics, paralinguistics, or contents in isolation. While these datasets are valuable, they rarely capture the contextual flow of dialogue or the reasoning process behind emotional states. 

Existing benchmarks, such as AudioBench~\cite{audiobench}, Dynamic-Superb~\cite{dynamicsuperb}, AIR-Bench~\cite{AIRbench}, OpenASQA~\cite{OpenASQA}, and MMAU~\cite{MMAU}, evaluate not only speech understanding but also paralinguistic tasks. However, their QA pairs are primarily derived from speech datasets focusing on isolated emotion and speaker-related tasks, without integration of contextual and paralinguistic cues within a unified QA format.

Contextual paralinguistic question answering (CPQA)~\cite{cpqa_interspeech} addresses this gap by requiring joint reasoning over contextual and paralinguistic information.
A CPQA dataset was created to evaluate empathetic reasoning in Speech-LLMs~\cite{cpqa_interspeech}, using a pipeline that condenses high-quality, emotion-rich speech data and leverages LLMs to automatically generate QA pairs. Nonetheless, this prior work neither validates the pipeline at scale nor assesses its effectiveness for model training.

Evaluating the CPQA task also presents its own challenges. LLMs are commonly employed as judges to assess Speech-LLM performance on tasks involving open-ended responses, such as contextual reasoning. However, a single evaluation prompt may not generalize well across the diverse range of question types found in QA tasks that incorporate both contextual and paralinguistic cues to varying extents.

To overcome these limitations, we propose a data-centric QA approach to build empathetic Speech-LLMs by combining explicit and implicit modeling of paralinguistic context:
\begin{itemize}
    \item Explicit Modeling: We inject structured paralinguistic metadata such as  emotion categories directly into model inputs during training, helping the model ground its responses in affective context.
\item Implicit Modeling: Building on prior work~\cite{cpqa_interspeech} that used only categorical emotion annotations, our approach enhances the QA generation pipeline by additionally incorporating dimensional emotion annotations (e.g., valence, arousal, dominance). This extension diversifies training data and aim to help the model better understand emotional nuances rather than discrete emotion categories. It also enables  the model to generalize effectively to unseen and complex emotional states. 


\end{itemize}

To better interpret model improvements, we investigate the reliability of LLM-based judge scores. While contextual reasoning lacks established alternatives to judge LLM scoring, the paralinguistic components of CPQA such as emotion understanding can be evaluated using standard classification metrics. 
Specifically, we assess whether a judge LLM can accurately infer classification-style outcomes and propose judge LLM evaluations with estimated accuracy and F1-score to enhance reliability for questions with deterministic answers.

This work lays a foundation for building emotionally intelligent speech-language systems that respond with both content relevance and empathetic awareness.




\begin{figure*}[h]
    \raggedright  
    \footnotesize
    \begin{tcolorbox}[colframe=black, colback=gray!10, arc=2mm, boxrule=0.5pt, width=1\textwidth]

        You are tasked with generating a set of paralinguistic questions and answers based on a given audio clip's characteristics. The QA pairs should serve as training data for multimodal large language models (LLMs) that rely exclusively on audio cues for reasoning. To achieve this, follow these instructions:

        \begin{enumerate}
            \item \textbf{Focus Areas:}
            \begin{itemize}
            \item Questions should explore speaker traits details such as emotions, gender, etc. and the reasoning behind emotional expressions and content in the audio.
            \item Use both **discrete emotion labels** and **continuous emotion dimensions** (arousal, dominance, valence) for emotion related QA.
            \item Ensure that both simple inquiries and complex reasoning queries are included.
                \item Combine information from the provided audio-derived emotion and gender labels along with the text transcription to generate QA pair. Note that emotion labels may not always be accurate, so analyze text also to refine your questions.

            \end{itemize}
            \item \textbf{Word-Level Metadata Guide}
            \begin{itemize}
                \item Each word is aligned with matched emotions and genders.
                \item For emotions: \textit{predict\_emo2vec} contains the discrete emotion label (e.g., `happy', `angry').
                \item \textit{predict\_dim} provides three scores in the order [arousal, dominance, valence]
            \end{itemize}

            \item \textbf{Diversity and Quality of Questions:}
            \begin{itemize}
                \item Craft a variety of question types that encourage comprehensive paralinguistic analysis of audio cues
            \end{itemize}

            \item \textbf{Question-Answer Types:}
            \begin{itemize}
                \item Do not reference any transcripts, text, or metadata labels in questions and answers. Just use the transcript and metadata (emotion and gender) to craft QA pairs. Avoid terms such as `text,' `transcript,' `metadata,' `label,' `timestamp,' `labeled,' etc. in both questions and answers.
                \item **Important Note:** Do *not* simply rephrase the example questions. Use them as a guide and apply your own analysis to generate QA pairs:
                \begin{itemize}
                    \item What is the primary emotion in the audio clip?
                 \item How does the speaker's emotion change over time?
                 \item What makes speaker \textit{emotion\_type} in this clip?
                 \item Why speaker is feeling \textit{emotion\_type} when mentioning \textit{situation}?
                 \item Why does the speaker become \textit{emotion\_type} in the end?
                 \item What is the gender of the speaker in this clip?

                \end{itemize}

            \item Do not generate one word answers. Be creative to generate answers like `speaker is female' or `speaker is feeling happy' for simple questions.
            \item Do not use model name or metadata file name in the question and answer text (for example, avoid phrases like `emotion predicted by emotion2vec' or `gender in the metadata file').
            \item Do not invent or hallucinate any data. Only use the provided word-level and paralinguistic metadata when answering the questions.
            \item Ensure that English usage is correct in the QA pairs.
            \end{itemize}             
    \end{enumerate}

    
    \textbf{Output Format:} Format each QA pair clearly with Q: and A: tags for the question and answer respectively.
    
    \textbf{Inputs:} Utterance: \texttt{`\{utterance\}`}, Paralinguistics data: \texttt{\{emotion\_gender\_level\_data\}}.
    \end{tcolorbox}
    \vspace{-2pt}
   \caption{Prompt for generating QA pairs from audio clips using both dimensional and categorical emotion annotations}
    \label{fig:prompt_qa_dim}
\end{figure*}

\section{Contextual-paralinguistic question answering}
\subsection{Paralinguistic question answering (PQA)}
Prior efforts to enhance Speech-LLMs for paralinguistic understanding have primarily relied on generating question–answer (QA) pairs from traditional speech corpora originally designed for isolated paralinguistic tasks, such as speaker identification, gender recognition, or emotion classification. A common practice involves using fixed question templates~\cite{audiobench}, where questions directly inquire about ground-truth paralinguistic labels (e.g., “What is the speaker's emotional state?” or “What is the speaker’s gender?”), and the corresponding label is used as the answer. We refer to this format as paralinguistic question answering (PQA).
Although some variability is introduced through multiple template variants, these questions are limited in both linguistic richness and contextual depth. Consequently, Speech-LLMs trained on such data often struggle to generalize beyond direct, label-oriented queries, particularly in scenarios requiring nuanced or context-dependent reasoning.

\subsection{Contextual-paralinguistic question answering (CPQA)}
In contrast to PQA, contextual-paralinguistic question answering (CPQA) integrates both contextual reasoning and paralinguistic understanding within a single question–answering task~\cite{cpqa_interspeech}. While many current Speech-LLMs are exposed to training data involving either contextual reasoning or isolated paralinguistic inference, they are rarely trained on tasks that require joint reasoning across both dimensions.
CPQA questions are designed such that answering them necessitates understanding contextual cues in tandem with paralinguistic signals. For instance, in the question “Why is the man angry?”, a Speech-LLM must localize segments of the speech that convey anger, determine the speaker's gender, and reason about the underlying cause based on the broader context. These multifaceted questions demand deeper audio-language integration and serve as a more realistic benchmark of empathetic and situational understanding in Speech-LLMs.

\subsection{Conventional CPQA data generation}
A recent approach for generating contextual paralinguistic question-answering (CPQA) data proposes a two-stage approach: (1) condensing emotion-rich speech data and (2) prompting large language models (LLMs) to generate QA pairs grounded in both the speech content and paralinguistic metadata~\cite{cpqa_interspeech}.
In the first stage, emotion and gender are estimated every 2 seconds of speech using dedicated recognition tools. Gender is predicted using a fine-tuned WavLM-ECAPA model~\cite{chen2022wavlm}. Emotion labels include categories from Emotion2Vec~\cite{emotion2vec} and dimensions from a continuous emotion recognition model~\cite{wagner2023dawn}. ASR tool WhisperX~\cite{Bain2023WhisperXTS} estimates the language of the speech samples, and obtain word-level transcription and time stamps.  After the meta data is obtained, a language filter filters the speech samples for the language of interest.  A speech emotion recognition (SER) consistency filter and an emotion occurrence filter ensures reliable emotion labels and obtain balanced emotion-rich speech corpora.
In the second stage, the aligned data containing word-level transcripts, emotion categories, and gender are used to prompt an LLM (GPT-4o) to generate CPQA pairs. 
Although emotion dimensions are estimated alongside categorical labels, they are only used in the SER consistency filter for data selection and are not leveraged during QA generation.

\subsection{Proposed CPQA data generation}
\label{ssec:proposed_data}
To generate more diverse QA, we propose an enhanced CPQA generation pipeline with two major improvements:
\begin{itemize}
    \item Inclusion of emotion dimensions: In addition to emotion categories and gender, we incorporate dimensional emotion annotations, valence, arousal, and dominance, into the QA generation process. These continuous signals offer complementary affective context and support more semantically nuanced questions, such as those exploring emotional intensity or ambiguity.
    \item Training-Scale CPQA Data using better prompt: We scale QA generation to create a large dataset suitable for directly training Speech-LLMs, enabling improved learning of contextual and paralinguistic reasoning. Compared to ~\cite{cpqa_interspeech}, we modify the prompting strategy (Fig.~\ref{fig:prompt_qa_dim}) to fully leverage these multi-dimensional paralinguistic cues during QA generation.
\end{itemize}

The use of both categorical and dimensional emotion representations allows for a richer supervision signal, facilitates generalization beyond predefined emotion labels, and supports flexible reasoning over subtle emotional states.



\section{Prompt with emotion metadata}
\label{sec:meta-prompt}
While training Speech-LLMs with CPQA data encourages the model to learn contextual-paralinguistic reasoning, it remains unclear how effectively the speech encoder extracts and conveys paralinguistic cues in a form interpretable by the LLM. To investigate this, we address two key questions: (1) What is the potential upper bound of model performance in CPQA with paralinguistic understanding? (2) Can explicitly provided paralinguistic metadata at inference compensate for a model lacking such capability? To explore these, we propose injecting structured paralinguistic metadata. In this work, we specifically on emotion cues and use time-stamped emotion labels as the metadata source. Question prompts are augmented as follows:
\begin{tcolorbox}[mylisting]
\begin{lstlisting}[basicstyle=\ttfamily\footnotesize]
 question = question + instruction1.replace("#XXXX#", emotion_labels) + instruction2
\end{lstlisting}
\end{tcolorbox}
where
\begin{tcolorbox}[mylisting]
\begin{lstlisting}[basicstyle=\ttfamily\footnotesize]
instruction1 = "If relevant, incorporate the following speech-derived emotion estimations (recorded every two seconds) when generating your answer: #XXXX#"
instruction2 = "All other time intervals without explicit emotion labels should be considered neutral. However, these emotion labels may not always be accurate. Analyze the content carefully and refine your response accordingly."
\end{lstlisting}
\end{tcolorbox}
An example of \texttt{emotion\_labels} is ``\texttt{2-4 second: sad, 10-12 second: angry, 12-14 second: angry.}''
We train Speech-LLMs using these augmented prompts and compare their performance with models trained on standard CPQA data without metadata. This setup allows us to estimate the performance upper bound achievable with perfect paralinguistic grounding.

Additionally, we apply the same metadata injection strategy at inference to evaluate whether explicitly provided emotional context can enhance models that have not been trained to extract such cues intrinsically. This helps us assess the extent to which external metadata can compensate for limited paralinguistic understanding.

\section{Interpretation of LLM judge score for classification }
\label{sec:estimated_acc}
Automatic evaluation of open-ended LLM responses often rely on other LLMs serving as judges.  However, for classification-type questions with definitive answers, such subjective evaluation is unnecessary. To interpret LLM judge's assessment of evaluation performance on these questions, we propose using estimated accuracy and F1-score, which are widely adopted in standard emotion and gender classification. We also investigate the correlation between these classification matrices and the scores given by the LLM judge. 

To compute accuracy and F1-score, we convert LLM-generated answers into classification labels using a two-step approach: (1) direct keyword matching, and (2) semantic similarity matching. This process enables reliable evaluation of tasks such as emotion classification and gender classification.
\begin{algorithm}[t]
\caption{Label Estimation from LLM-Generated Answer}
\label{algo:algo1}
\begin{algorithmic}[1]
\REQUIRE Answer text $a_P$, label set $L = \{l_1, l_2, ..., l_n\}$, embedding extraction function $f(\cdot)$
\STATE Compute answer embedding $e_P = f(a_P)$
\STATE Compute label embeddings $\{e_1, e_2, ..., e_n\}$ where $e_i = f(l_i)$ for each $l_i \in L$
\vspace{0.5em}\\
\COMMENT{Step 1: Keyword Matching}
\FOR{each label $l_i \in L$}
    \IF{$l_i$ appears in $a_P$}
        \STATE \textbf{return} $l_i$
    \ENDIF
\ENDFOR
\vspace{0.5em}\\
\COMMENT{Step 2: Semantic Similarity Matching}
\FOR{each $e_i$ in label embeddings}
    \STATE Compute similarity $s_i = \cos(e_i, e_P)$
\ENDFOR
\STATE $\hat{l} \leftarrow l_{\arg\max_{i \in 1..n} s_i}$

\STATE \textbf{return} $\hat{l}$
\end{algorithmic}
\end{algorithm}

If no keyword matches, we assign the label with the highest cosine similarity between the predicted answer's embedding and each class label's embedding:
\begin{equation}
    \hat{l} = \arg\max_{l_i\in L} \text{cos}(f(l_i), f(a_P))     
    \end{equation}
where $L$ is the set of label embeddings, and $a_P$ is the embedding of the LLM-generated answer. $f(\cdot)$ is the embedding extraction function. We use the \textit{paraphrase-MiniLM-L6-v2}\footnote{\url{https://huggingface.co/sentence-transformers/paraphrase-MiniLM-L6-v2}} model from SentenceTransformers due to its effectiveness in semantic similarity tasks ~\cite{reimers-2019-sentence-bert}.  The complete procedure is outlined in Algorithm~\ref{algo:algo1}.

\section{Experiments}
\label{sec:exp}
\subsection{Experimental setting}
\subsubsection{Network structures}
We follow the MERaLiON Audio-LLM framework~\cite{meralion}\footnote{\url{https://huggingface.co/MERaLiON}}, which includes a speech encoder, a text decoder, and an adapter that bridges the modality gap by aligning the hidden dimensions.
In our setup, we adopt the Whisper large-v3 encoder\footnote{\url{https://huggingface.co/openai/whisper-large-v3}}~\cite{radford2022whisper} which outputs sequences of length 1,500 with a hidden size of 1,280.
To interface with the decoder, a lightweight multi-layer perceptron (MLP) adapter with two hidden layers compresses and transforms the encoder output into 100 speech token embeddings of dimension 3,854, matching the decoder's input space.
We adopt the Gemma-2 9B Instruct model\footnote{\url{https://huggingface.co/google/gemma-2-9b-it}}~\cite{gemma_2024} as the text decoder. It processes a concatenation of speech token embeddings and text instruction to generate natural language responses. During training, both the encoder and decoder are frozen, and only the adapter is updated. We fix the number of steps of 120,000 for fair comparison between models. The learning rate is set to $10^{-4}$.


\begin{table}[t]
\centering
\caption{Statistics of Training Datasets. ``*'' is to differentiate the proposed PQA* set from the conventional PQA sets.}
\begin{tabular}{c|l|l|r}
\hline
 & Dataset &Corpora& QA pairs \\
\hline
\multirow{10}{*}{Baseline} &\multirow{4}{*}{ASR} 
  & IMDA Part 3         & 119,888 \\
 & & IMDA Part 6         & 141,480 \\
 & & LibriSpeech clean-100  & 104,014 \\
 & & LibriSpeech clean-360  & 28,539 \\
 & & LibriSpeech others-500  & 148,688 \\
  \cline{3-4}
&  & Total     & 542,609 \\
\cline{2-4}
&\multirow{3}{*}{Gender-PQA} 
  & IEMOCAP         & 9,035 \\
 & & VoxCeleb        & 148,642 \\
 & & IMDA Part 3     & 119,685 \\
  \cline{3-4}
 & & Total     & 277,362 \\
\cline{2-4}
&\multirow{3}{*}{Emotion-PQA} 
  & IEMOCAP             & 9,035 \\
  && MSP-Podcast Train            & 84,030 \\
  && In-house data 1         & 125,983 \\
  \cline{3-4}
  && Total     & 219,048 \\
\hline
\multirow{3}{*}{Proposed} &  PQA*
  & MSP-Podcast Train             & 443,815 \\
  \cline{2-4}
  & CPQA &    In-house data 2            & 32,960 \\
\hline
\end{tabular}
\label{tab:dataset_stats}
\end{table}

\subsubsection{Datasets}

We construct our training data from question–answering (QA) datasets derived from automatic speech recognition (ASR), gender recognition (GR), and emotion recognition (ER) corpora. The ASR datasets are included so that the model learns linguistic content in speech and also shown to improve emotion recognition performance \cite{kang2024frozen}. The detailed statistics of QA pairs generated from each dataset as well as the task wise are shown in Table~\ref{tab:dataset_stats}.

The ASR data sets include the IMDA dataset Part 3 and Part 6~\cite{wang2025advancing,koh19_interspeech} that are conversational and spontaneous speech, as well as LibriSpeech's clean-100, clean-360, and other-500 subsets~\cite{librispeech}.
The GR corpora includes the IEMOCAP~\cite{iemocap}, VoxCeleb1~\cite{VoxCeleb1}, and IMDA Part 3 datasets.
The ER corpora include IEMOCAP~\cite{iemocap}, the MSP-Podcast Train set~\cite{msp}, and an in-house dataset of movies and TVs consisting of 125,983 speech samples and annotations of emotion category  of \textit{angry, disgusted, fearful, happy, sad, surprised, embarrassment, sarcasm}, and \textit{worry}. For these QA generation, QA templates are used~\cite{audiobench}. These datasets are used to train the baseline system. 

The proposed PQA* and CPQA datasets are generated using the GPT-4o API (Azure version 2024-07-01-preview)\footnote{\url{https://learn.microsoft.com/en-us/azure/ai-services/openai/}} (see Section~\ref{ssec:proposed_data}).  The PQA* training set is created from the MSP-Podcast training set using similar prompt as shown in Fig.~\ref{fig:prompt_qa_dim} except instruction for contextual questions. For the CPQA task, we first curate a balanced emotion-rich subset of 4,740 speech clips (20–30 seconds each) from an in-house dataset, following the data condensation framework in~\cite{cpqa_interspeech}. We use valence ranges of $[0,0.5)$, $(0.5,1.0]$ and $[0.4, 0.6]$ for the negative, positive, and neutral categories, respectively, in the SER consistency filter that ensures annotation reliability. Lower thresholds are used for the emotion occurrence filter to collect a larger dataset, with minimum counts set to $[3, 3, 2, 2, 1, 1]$ for the categories \textit{angry}, \textit{happy}, \textit{sad}, \textit{surprised}, \textit{disgusted}, and \textit{fearful}, respectively. Using this data, we generate the CPQA dataset of 32,960 QA pairs following (see Section~\ref{ssec:proposed_data}). 

We evaluate model performance using both human-annotated and automatically generated CPQA datasets\footnote{https://huggingface.co/datasets/MERaLiON/CPQA-Evaluation-Set}.  First, we collect 480 speech clips (10–30 seconds each) using the same data condensation pipeline as for training.  Three human annotators listen to each clip, correct estimated emotion and gender labels, and generate QA pairs designed to assess contextual-paralinguistic understanding, with a focus on emotion and gender. Each QA pair is categorized into one of the following: contextual only (C), contextual with emotion (CE), or contextual with gender (CG), resulting in 70, 303, and 88 QA pairs, respectively.
Due to the annotation workload, we supplement this set with an automatically generated CPQA evaluation set using the same GPT-4o API using only emotion category metadata (excluding emotion dimensions),  following the validation approach from~\cite{cpqa_interspeech}. This LLM-generated set includes 3,396 QA pairs.
Additionally, we evaluate on two emotion-PQA benchmarks: the IEMOCAP test set~\cite{iemocap} from AudioBench~\cite{audiobench} and a constructed QA set from the MSP-Podcast Test Set 2~\cite{msp} following the same QA template. System configurations are shown in Table~\ref{tab:systems}.

\subsubsection{Evaluation metric}
we employ AudioBench~\cite{audiobench}
for the assessment that uses gpt4o-as-judge to evaluate task performance. Each response is scored on a scale from 0 to 5, based on criteria such as relevance, coherence, and accuracy. The scores are linearly rescaled to a 0–100 range for interpretability. For the emotion-PQA, we further use estimated weighted accuracy and F1-score proposed in Section~\ref{sec:estimated_acc}. 

\begin{table}[t]
\centering
\caption{System configurations for training and inference.}
\vspace{-5pt}
\begin{tabular}{l l c c}
\toprule
\multirow{2}{*}{System} & \multirow{2}{*}{Training Data} & \multicolumn{2}{c}{Emotion labels in prompts}\\
\cline{3-4}
& & Train & Inference \\
\midrule
S10 & Baseline & \xmark & \xmark \\
S11 & Baseline & \xmark & \cmark \\
S20 & Baseline + PQA* & \xmark & \xmark \\
S21 & Baseline + PQA* & \xmark & \cmark \\
S30 & Baseline + PQA* + CPQA & \xmark & \xmark \\
S31 & Baseline + PQA* + CPQA & \xmark & \cmark \\
S32 & Baseline + PQA* + CPQA & \cmark & \cmark \\
\bottomrule
\vspace{-10pt}
\label{tab:systems}
\end{tabular}
\end{table}

\subsection{Experimental results and analysis}
\subsubsection{Training using the proposed LLM-generated QA sets}
\begin{table}[t]
\caption{Performance on CPQA and PQA tasks. ``Human'' and ``LLM'' refer to human-annotated and LLM-generated CPQA evaluation sets, respectively.}
\begin{center}
\vspace{-5pt}
\begin{tabular}{l|cc|cc}
\hline
System &\multicolumn{2}{c|}{CPQA} & \multicolumn{2}{c}{PQA} \\
 & Human & LLM & IEMOCAP & MSP-Podcast\\
\hline
S10       &41.00 & 41.50 & 50.76 & 46.34 \\
S20  & 52.06 & 51.51 & 46.63 &  53.31 \\
S30          & 56.75  & 58.89
 & 45.70 &54.36 \\
\hline
\end{tabular}
\label{tab:train_data}
\end{center}
\end{table}

\begin{table}[t]
\caption{Question-type-wise analysis of the human-annotated CPQA evaluation set.}
\vspace{-5pt}
\begin{center}
\begin{tabular}{l|ccc}
\hline
System & C & CE & CG  \\
\hline
S10        & 44.86 & 37.56 & 49.77 \\
S20   & 54.57 & 48.98 & 60.68 \\
S30         & 60.29&	53.86	&63.86 
 \\
\hline
\end{tabular}
\label{tab:type-wise}
\end{center}
\end{table}
\vspace{-5pt}
We evaluate the proposed data generation methods by comparing models trained on the baseline dataset (S10) with those additionally trained on the proposed PQA* and CPQA data sets. As shown in Table~\ref{tab:train_data}, S20  (baseline + PQA*) significantly outperforms S10 on contextual paralinguistic QA tasks in both human-annotated and LLM-generated evaluation sets. S30, which incorporates both proposed datasets, achieves the best overall performance, with score improvements of 38.41\% and 41.90\% over S10. For PQA evaluation, we observe a performance drop on IEMOCAP but a slight gain on MSP-Podcast. Since the PQA* training set in S20 is generated from the MSP-Podcast training set, this likely amplifies domain mismatch when evaluated on IEMOCAP. Additionally, the CPQA data used in S30 are generated using emotion labels estimated by speech emotion recognition (SER) tools, which may introduce noise and reduce accuracy in direct emotion-based PQA tasks.

\subsubsection{Question-type-wise analysis in CPQA evaluation }
We analyze the impact of the generated training data across question types in the human-annotated CPQA set, which includes type labels: contextual questions only (C), contextual + emotion (CE), and contextual + gender (CG). Table~\ref{tab:type-wise} shows that performance is lowest on CE questions, highlighting their difficulty. Adding the proposed PQA* and CPQA training data improves scores on CE and CG questions by 43.40\% and 28.31\%, respectively. Notably, it also improves performance on contextual-only questions by 34.40\%, indicating that proposed training data generation  enhances contextual understanding, not just  paralinguistic reasoning.

\subsubsection{Emotion metadata in prompts}
\begin{figure}[t]
\vspace{-5pt}
    \centering    \includegraphics[width=0.48\textwidth]{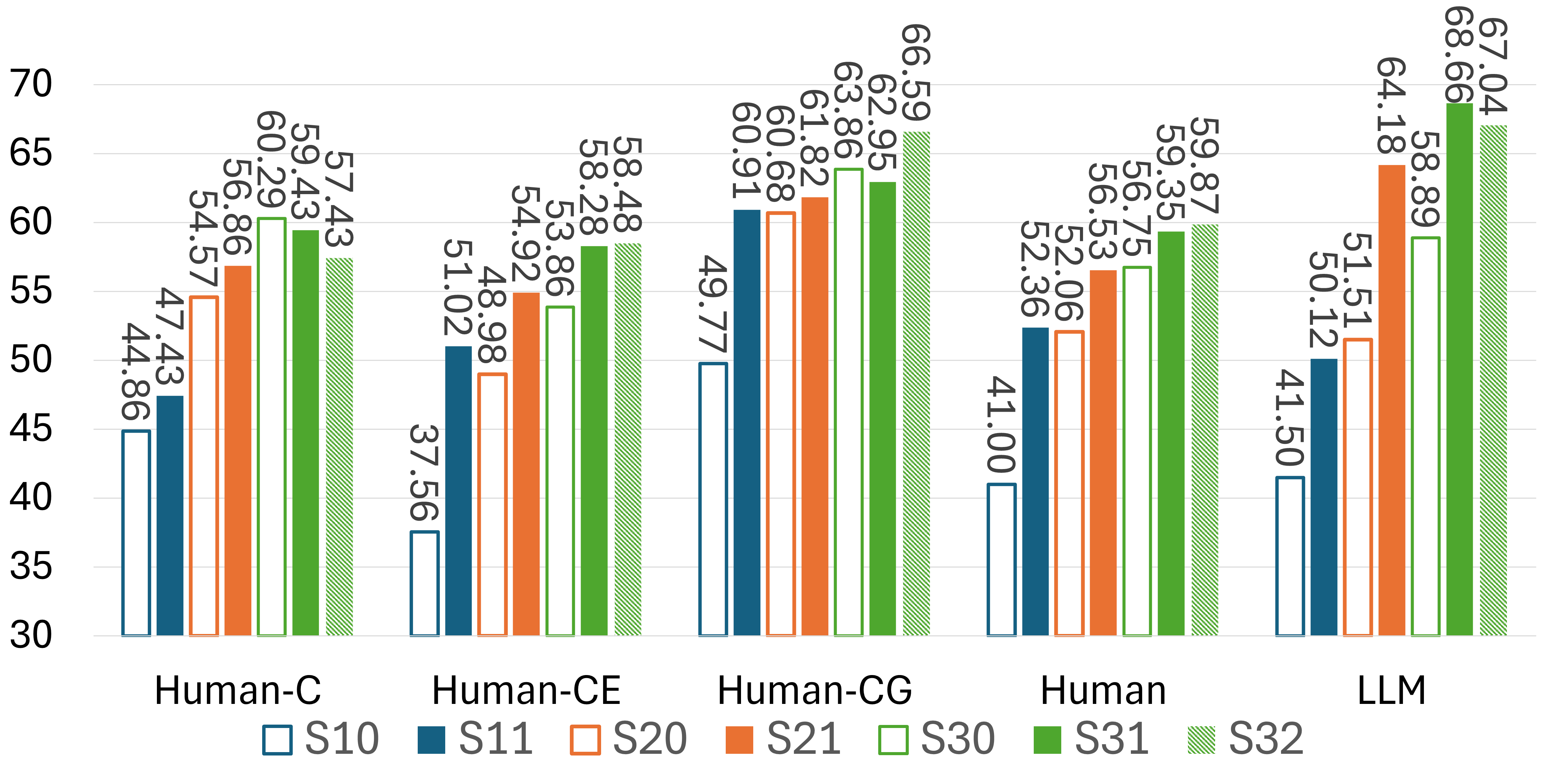}
    \vspace{-10pt}
    \caption{Impact of emotion metadata in training and inference prompt. Performance breakdown by question type (C, CE, CG) within the human-annotated CPQA set, along with the weighted average score for the full human-annotated set and performance on the LLM-generated CPQA set.}
    \label{fig:ser_in_prompt}
    \vspace{-8pt}
\end{figure}

We next  investigate the effect of explicitly adding emotion metadata, specifically time-stamped emotion labels,  in the question prompts during training and inference. For the human-annotated set, as shown in Fig.~\ref{fig:ser_in_prompt},  adding such emotion metadata at inference only (S10 vs S11) for the baseline model results in substantial performance gains across all question types in the human-annotated set, especially for contextual+emotion (CE) questions  (+35.84\%). It indicates that explicitly provided paralinguistic metadata at inference compensate for a model
lacking such capability. 
When training includes the proposed PQA* set, the gains from adding emotion metadata in inference (S20 vs. S21) are smaller. It suggests the PQA* set improves implicit learning of paralinguistic cues.  

Model  trained with CPQA data (S30) outperform others even those with the emotion metadata in inference, confirming that training with contextual-paralinguistic data is more effective than relying solely on explicit emotion meta data during inference.
 This also may show that the implicit embedded emotion information in CPQA data in training is less affected by noisy emotion labels compared to explicitly providing in inference prompts.  
Adding emotion metadata at inference on top of CPQA training (S30 vs. S31) shows mixed results: a slight drop for contextual-only (C) and contextual+gender (CG) questions, likely due to irrelevant or conflicting metadata, but a small improvement for CE questions, suggesting that explicit cues can still complement learned representations.

Finally, when emotion metadata is explicitly provided in the CPQA prompt in both training and inference stages (S32), CE performance reaches its highest score, and CG performance remains strong. This setting may represent a potential upper bound for CE questions, assuming the emotion metadata are sufficiently accurate to replace ground-truth labels. However, C performance continues to decline. These results indicate that explicit emotion metadata is highly beneficial for CE questions, especially when aligned across training and inference, while a trade-off exists between general contextual understanding and integration with paralinguistic cue. Explicit emotion metadata injection may not benefit all tasks, whereas the implicit method, where the model learns to integrate emotional information, may offer better generalization across diverse question types. Overall, the human-annotated set achieves its best performance at score of 59.87 with 46.03\% increase compared to S10.

Compared to human-annotated set, we observed greater performance gains on the LLM-generated evaluation set when emotion metadata was added in the inference prompt. This may be due to the unintended inclusion of direct emotion questions, despite instructions to avoid them during QA generation. Incorporating time-stamped emotion cues in such questions can inadvertently reveal the answer, compromising the validity of the evaluation. Stricter controls are needed in future QA generation to ensure robust and unbiased assessments. 

Emotion metadata used in training and inference is derived from SER models rather than ground-truth labels. While prompts clarify this to the LLM, inaccuracies may still affect performance. Nonetheless, this explicit approach enables scalable analysis and reveals how Speech-LLMs leverage contextual and paralinguistic cues.
Explicitly including emotion metadata in prompts not only approximates an upper-bound scenario, but also represents a valid approach, simulating a pre-processing SER module for enhanced model input.

\subsubsection{Interpret LLM's answers in classification tasks}
\begin{table}[t]
\caption{Analysis of direct emotion questions using estimated weighted acc* and F1*, as well as scores by the LLM judge.}
\vspace{-5pt}
\begin{center}
\begin{tabular}{l|ccc|ccc}
\hline
 &\multicolumn{3}{c|}{\textbf{IEMOCAP}}  &\multicolumn{3}{c}{\textbf{MSP-Podcast}}  
\\
System    & LLM & acc* & F1* & LLM & acc* & F1* \\
\hline
S10            & 50.76 & 42.03 & 41.90    
                    & 46.34 & 40.37 & 39.31    
                    \\
S20       & 46.63 & 33.76 & 33.92
                    & 53.31 & 37.54 & 36.52    
                    \\
                    
S30   & 45.70 & 33.17 & 33.43

                & 54.36 & 40.01 & 37.25
                    \\
\hline
\end{tabular}

\label{tab:de}
\end{center}
\end{table}

\begin{figure}[t]
\vspace{-5pt}
    \centering    \includegraphics[width=0.48\textwidth]{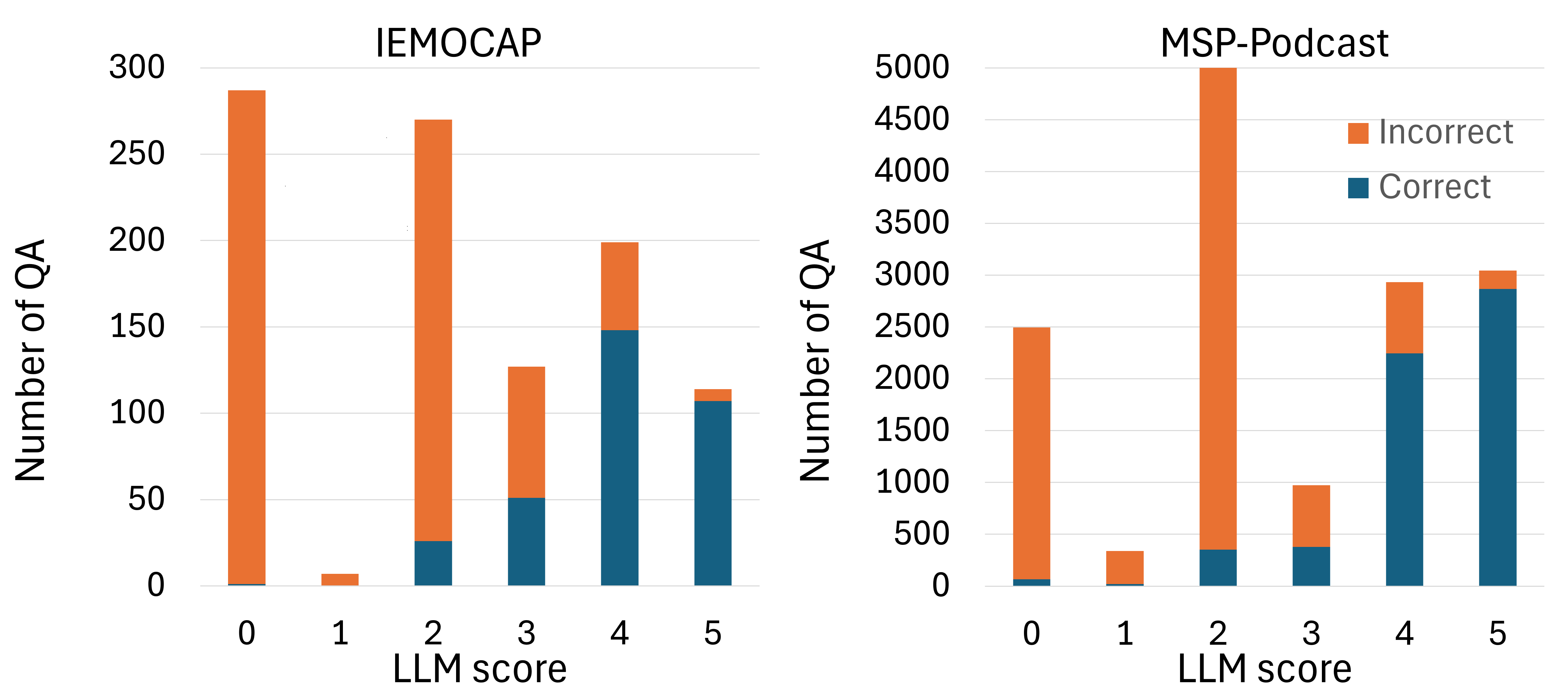}
    \vspace{-5pt}
    \caption{Distribution of original LLM scores (before scaling to 0--100) with system S30 alongside the proportions of correct and incorrect predictions within each score group. ``Correct'' and ``incorrect'' indicate whether the estimated labels match the ground truths.}
    \label{fig:de_freq}
\vspace{-5pt}
\end{figure}

We assess the trained Speech-LLMs on the direct emotion-PQA datasets from IEMOCAP and MSP-Podcast corpora using the proposed estimated weighted accuracy and F1-scores, as shown in Table~\ref{tab:de}. We focus on the comparison between metrics rather than those between the models that is discussed in Table~\ref{tab:train_data}. We observe a consistent correlation between LLM scores and the estimated accuracy and F1-scores. 
Note that we used all emotion categories 8 for IEMOCAP and 9 for MSP-Podcast in our evaluation. The observed accuracies, around 40\%, are substantially higher than random chance. 

To further illustrate this relationship, Fig.~\ref{fig:de_freq} shows the distribution of LLM scores alongside the proportions of correct and incorrect predictions within each score group. Although each prediction is either correct or incorrect, the LLM score distribution is not strictly bimodal. Instead, a significant number of predictions receive mid-range scores (2, 3, and 4), highlighting the limitations of using a single LLM scoring system for both classification-type and open-ended questions. Nonetheless, the ratio of correct to incorrect answers increases steadily with higher LLM scores, further validating the correlation between LLM scores and actual classification accuracy.

\section{Conclusion}
This work proposes methods to improve Speech-LLMs in empathetic reasoning by incorporating contextual and paralinguistic information: explicit modeling, which injects structured metadata during training, and implicit modeling, which uses a novel QA dataset generated from paralinguistic annotations and transcriptions. While explicit emotion metadata at inference can compensate for limited emotional understanding, and its use in both training and inference achieves the best performance on emotion-related QA, the implicit approach, which trains on contextual-paralinguistic data, proves more effective and generalized across diverse questions. We also propose classification-based metrics to validate LLM judges, offering an alternative comprehensive evaluation framework. 




\bibliographystyle{IEEEbib}
\bibliography{custom}

\begin{thebibliography}{10}

\bibitem{qwen}
Yunfei Chu, Jin Xu, Xiaohuan Zhou, Qian Yang, Shiliang Zhang, Zhijie Yan, Chang Zhou, and Jingren Zhou,
\newblock ``Qwen-audio: Advancing universal audio understanding via unified large-scale audio-language models,''
\newblock {\em arXiv preprint arXiv:2311.07919}, 2023.

\bibitem{qwen2}
Yunfei Chu, Jin Xu, Qian Yang, Haojie Wei, Xipin Wei, Zhifang Guo, Yichong Leng, Yuanjun Lv, Jinzheng He, Junyang Lin, Chang Zhou, and Jingren Zhou,
\newblock ``Qwen2-audio technical report,''
\newblock {\em arXiv preprint arXiv:2407.10759}, 2024.

\bibitem{gpt4}
Josh Achiama, Steven Adler, Sandhini Agarwal, Lama Ahmad, Ilge Akkaya, Florencia~Leoni Aleman, Diogo Almeida, Janko Altenschmidt, Sam Altman, Shyamal Anadkat, Red Avila, Igor Babuschkin, Suchir Balaji, and other,
\newblock ``{GPT}-4 technical report,''
\newblock {\em arXiv preprint arXiv:2303.08774}, 2023.

\bibitem{salmonn}
Changli Tang, Wenyi Yu, Guangzhi Sun, Xianzhao Chen, Tian Tan, Wei Li, Lu~Lu, Zejun Ma, and Chao Zhang,
\newblock ``Salmonn: Towards generic hearing abilities for large language models,''
\newblock {\em arXiv preprint arXiv:2310.13289}, 2023.

\bibitem{Lin24}
Guan-Ting Lin, Cheng-Han Chiang, and Hung-Yi Lee,
\newblock ``Advancing large language models to capture varied speaking styles and respond properly in spoken conversations,''
\newblock {\em arXiv preprint arXiv:2402.12786}, 2024.

\bibitem{kimparalinguistics}
Heeseung Kim, Soonshin Seo, Kyeongseok Jeong, Ohsung Kwon, Soyoon Kim, Jungwhan Kim, Jaehong Lee, Eunwoo Song, Myungwoo Oh, Jung-Woo Ha, Sungroh Yoon, and Kang~Min Yoo,
\newblock ``Paralinguistics-aware speech-empowered large language models for natural conversation,''
\newblock in {\em The Thirty-eighth Annual Conference on Neural Information Processing Systems (NeurIPS)}, 2024.

\bibitem{kang2024frozen}
Wonjune Kang, Junteng Jia, Chunyang Wu, Wei Zhou, Egor Lakomkin, Yashesh Gaur, Leda Sari, Suyoun Kim, Ke~Li, Jay Mahadeokar, et~al.,
\newblock ``Frozen large language models can perceive paralinguistic aspects of speech,''
\newblock {\em arXiv preprint arXiv:2410.01162}, 2024.

\bibitem{wang24blsp}
Chen Wang, Minpeng Liao, Zhongqiang Huang, Junhong Wu, Chengqing Zong, and Jiajun Zhang,
\newblock ``{BLSP-Emo}: Towards empathetic large speech-language models,''
\newblock {\em Proceedings of the 2024 Conference on Empirical Methods in Natural Language Processing (EMNLP)}, 2024.

\bibitem{desta}
Ke-Han Lu, Zhehuai Chen, Szu-Wei Fu, He~Huang, Boris Ginsburg, Yu-Chiang Wang, and Hung-yi Lee,
\newblock ``{DeSTA}: Enhancing speech language models through descriptive speech-text alignment,''
\newblock in {\em Interspeech}, 2024, pp. 4159--4163.

\bibitem{desta2}
Ke-Han Lu, Zhehuai Chen, Szu-Wei Fu, Chao-Han~Huck Yang, Jagadeesh Balam, Boris Ginsburg, Yu-Chiang Wang, and Hung-yi Lee,
\newblock ``{DeSTA2}: Developing instruction-following speech language model without speech instruction-tuning data,''
\newblock in {\em IEEE International Conference on Acoustics, Speech and Signal Processing (ICASSP)}, 2025.

\bibitem{wu-etal-2025-beyond}
Zehui Wu, Ziwei Gong, Lin Ai, Pengyuan Shi, Kaan Donbekci, and Julia Hirschberg,
\newblock ``Beyond silent letters: Amplifying {LLM}s in emotion recognition with vocal nuances,''
\newblock in {\em Findings of the Association for Computational Linguistics: NAACL 2025}, 2025, pp. 2202--2218.

\bibitem{CLAP4Emo}
Wei-Cheng Lin, Shabnam Ghaffarzadegan, Luca Bondi, Abinaya Kumar, Samarjit Das, and Ho-Hsiang Wu,
\newblock ``{CLAP4Emo: ChatGPT-Assisted Speech Emotion Retrieval with Natural Language Supervision},''
\newblock in {\em IEEE International Conference on Acoustics, Speech and Signal Processing (ICASSP)}, 2024, pp. 11791--11795.

\bibitem{xu2024secap}
Yaoxun Xu, Hangting Chen, Jianwei Yu, Qiaochu Huang, Zhiyong Wu, Shi-Xiong Zhang, Guangzhi Li, Yi~Luo, and Rongzhi Gu,
\newblock ``{SEC}ap: Speech emotion captioning with large language model,''
\newblock in {\em Proceedings of the AAAI Conference on Artificial Intelligence}, 2024, vol.~38, pp. 19323--19331.

\bibitem{wu2024empower}
Haibin Wu, Huang-Cheng Chou, Kai-Wei Chang, Lucas Goncalves, Jiawei Du, Jyh-Shing~Roger Jang, Chi-Chun Lee, and Hung-Yi Lee,
\newblock ``Empower typed descriptions by large language models for speech emotion recognition,''
\newblock in {\em Asia Pacific Signal and Information Processing Association Annual Summit and Conference (APSIPA ASC)}. IEEE, 2024, pp. 1--6.

\bibitem{cheng2025voxdialogue}
Xize Cheng, Ruofan Hu, Xiaoda Yang, Jingyu Lu, Dongjie Fu, Zehan Wang, Shengpeng Ji, Rongjie Huang, Boyang Zhang, Tao Jin, et~al.,
\newblock ``{VoxDialogue}: Can spoken dialogue systems understand information beyond words?,''
\newblock in {\em International Conference on Learning Representations (ICLR)}, 2025.

\bibitem{pandey2025sift}
Prabhat Pandey, Rupak~Vignesh Swaminathan, KV~Girish, Arunasish Sen, Jian Xie, Grant~P Strimel, and Andreas Schwarz,
\newblock ``{SIFT-50M}: A large-scale multilingual dataset for speech instruction fine-tuning,''
\newblock {\em arXiv preprint arXiv:2504.09081}, 2025.

\bibitem{gong2023joint}
Yuan Gong, Alexander~H Liu, Hongyin Luo, Leonid Karlinsky, and James Glass,
\newblock ``Joint audio and speech understanding,''
\newblock in {\em IEEE Automatic Speech Recognition and Understanding Workshop (ASRU)}. IEEE, 2023, pp. 1--8.

\bibitem{audiobench}
Bin Wang, Xunlong Zou, Geyu Lin, Shuo Sun, Zhuohan Liu, Wenyu Zhang, Zhengyuan Liu, AiTi Aw, and Nancy~F Chen,
\newblock ``Audiobench: A universal benchmark for audio large language models,''
\newblock {\em Proceedings of the 2025 Conference of the Nations of the Americas Chapter of the Association for Computational Linguistics (NAACL)}, pp. 4297--4316, 2025.

\bibitem{dynamicsuperb}
Chien-Yu Huang, Ke-Han Lu, Shih-Heng Wang, Chi-Yuan Hsiao, Chun-Yi Kuan, Haibin Wu, Siddhant Arora, Kai-Wei Chang, Jiatong Shi, Yifan Peng, Roshan Sharma, Shinji Watanabe, Bhiksha Ramakrishnan, Shady Shehata, and Hung-Yi Lee,
\newblock ``Dynamic-superb: Towards a dynamic, collaborative, and comprehensive instruction-tuning benchmark for speech,''
\newblock in {\em IEEE International Conference on Acoustics, Speech and Signal Processing (ICASSP)}, pp. 12136--12140.

\bibitem{AIRbench}
Qian Yang, Jin Xu, Wenrui Liu, Yunfei Chu, Ziyue Jiang, Xiaohuan Zhou, Yichong Leng, Yuanjun Lv, Zhou Zhao, Chang Zhou, et~al.,
\newblock ``{AIR-Bench}: Benchmarking large audio-language models via generative comprehension,''
\newblock {\em Proceedings of the 62nd Annual Meeting of the Association for Computational Linguistics}, pp. 1979--1998, 2024.

\bibitem{OpenASQA}
Yuan Gong, Hongyin Luo, Alexander~H Liu, Leonid Karlinsky, and James Glass,
\newblock ``Listen, think, and understand,''
\newblock in {\em International Conference on Learning Representations (ICLR)}, 2024.

\bibitem{MMAU}
S~Sakshi, Utkarsh Tyagi, Sonal Kumar, Ashish Seth, Ramaneswaran Selvakumar, Oriol Nieto, Ramani Duraiswami, Sreyan Ghosh, and Dinesh Manocha,
\newblock ``{MMAU}: A massive multi-task audio understanding and reasoning benchmark,''
\newblock in {\em International Conference on Learning Representations (ICLR)}, 2025.

\bibitem{cpqa_interspeech}
Qiongqiong Wang, Hardik~B Sailor, Tianchi Liu, and Ai~Ti Aw,
\newblock ``Contextual paralinguistic data creation for multi-modal {Speech-LLM}: Data condensation and spoken {QA} generation,''
\newblock in {\em Proc. Interspeech}, 2025.

\bibitem{chen2022wavlm}
Sanyuan Chen, Chengyi Wang, Zhengyang Chen, Yu~Wu, Shujie Liu, Zhuo Chen, Jinyu Li, Naoyuki Kanda, Takuya Yoshioka, Xiong Xiao, et~al.,
\newblock ``{WavLM}: Large-scale self-supervised pre-training for full stack speech processing,''
\newblock {\em IEEE Journal of Selected Topics in Signal Processing}, vol. 16, no. 6, pp. 1505--1518, 2022.

\bibitem{emotion2vec}
Ziyang Ma, Zhisheng Zheng, Jiaxin Ye, Jinchao Li, Zhifu Gao, Shiliang Zhang, and Xie Chen,
\newblock ``emotion2vec: Self-supervised pre-training for speech emotion representation,''
\newblock {\em Findings of the Association for Computational Linguistics (ACL)}, 2024.

\bibitem{wagner2023dawn}
Johannes Wagner, Andreas Triantafyllopoulos, Hagen Wierstorf, Maximilian Schmitt, Felix Burkhardt, Florian Eyben, and Bj{\"o}rn~W Schuller,
\newblock ``Dawn of the transformer era in speech emotion recognition: Closing the valence gap,''
\newblock {\em IEEE Transactions on Pattern Analysis and Machine Intelligence}, pp. 1--13, 2023.

\bibitem{Bain2023WhisperXTS}
Max Bain, Jaesung Huh, Tengda Han, and Andrew Zisserman,
\newblock ``{WhisperX}: Time-accurate speech transcription of long-form audio,''
\newblock in {\em Proc. Interspeech}, 2023.

\bibitem{reimers-2019-sentence-bert}
Nils Reimers and Iryna Gurevych,
\newblock ``Sentence-bert: Sentence embeddings using siamese bert-networks,''
\newblock in {\em Proceedings of the 2019 Conference on Empirical Methods in Natural Language Processing (EMNLP)}. 11 2019, Association for Computational Linguistics.

\bibitem{meralion}
Yingxu He, Zhuohan Liu, Shuo Sun, Bin Wang, Wenyu Zhang, Xunlong Zou, Nancy~F Chen, and Ai~Ti Aw,
\newblock ``{MERaLiON-AudioLLM}: Technical report,''
\newblock {\em arXiv preprint arXiv:2412.09818}, 2024.

\bibitem{radford2022whisper}
Alec Radford, Jong~Wook Kim, Tao Xu, Greg Brockman, Christine McLeavey, and Ilya Sutskever,
\newblock ``Robust speech recognition via large-scale weak supervision,''
\newblock {\em Proceedings of the 40th International Conference on Machine Learning (ICML)}, 2023.

\bibitem{gemma_2024}
Gemma Team, Thomas Mesnard, Cassidy Hardin, Robert Dadashi, Surya Bhupatiraju, Shreya Pathak, Laurent Sifre, Morgane Rivi{\`e}re, Mihir~Sanjay Kale, Juliette Love, Pouya Tafti, and Hussenot~Léonard others,
\newblock ``Gemma: Open models based on gemini research and technology,''
\newblock {\em arXiv preprint arXiv:2403.08295}, 2024.

\bibitem{wang2025advancing}
Bin Wang, Xunlong Zou, Shuo Sun, Wenyu Zhang, Yingxu He, Zhuohan Liu, Chengwei Wei, Nancy~F Chen, and AiTi Aw,
\newblock ``Advancing {Singlish} understanding: Bridging the gap with datasets and multimodal models,''
\newblock {\em arXiv preprint arXiv:2501.01034}, 2025.

\bibitem{koh19_interspeech}
Jia~Xin Koh, Aqilah Mislan, Kevin Khoo, Brian Ang, Wilson Ang, Charmaine Ng, and Ying-Ying Tan,
\newblock ``Building the {Singapore English} national speech corpus,''
\newblock in {\em Proc. Interspeech}, 2019, pp. 321--325.

\bibitem{librispeech}
Vassil Panayotov, Guoguo Chen, Daniel Povey, and Sanjeev Khudanpur,
\newblock ``Librispeech: An {ASR} corpus based on public domain audio books,''
\newblock in {\em IEEE International Conference on Acoustics, Speech and Signal Processing (ICASSP)}, 2015, pp. 5206--5210.

\bibitem{iemocap}
Carlos Busso, Murtaza Bulut, Chi-Chun Lee, Ebrahim~(Abe) Kazemzadeh, Emily~Mower Provost, Samuel Kim, Jeannette~N. Chang, Sungbok Lee, and Shrikanth~S. Narayanan,
\newblock ``{IEMOCAP}: interactive emotional dyadic motion capture database,''
\newblock {\em Language Resources and Evaluation}, vol. 42, pp. 335--359, 2008.

\bibitem{VoxCeleb1}
Arsha Nagrani, Joon~Son Chung, and Andrew Zisserman,
\newblock ``Voxceleb: A large-scale speaker identification dataset,''
\newblock in {\em Proc. Interspeech}, 2017, pp. 2616--2620.

\bibitem{msp}
Carlos Busso, Siddharth Narayanan, Emily Mower~Provost, Yue Zhang, Asterios Matsoukas, and Najim Dehak,
\newblock ``The {MSP-Podcast} corpus for speech emotion recognition,''
\newblock {\em IEEE Transactions on Affective Computing}, 2023.

\end{thebibliography}

\end{document}